\documentclass[sigconf]{acmart}
\AtBeginDocument{%
  }

\copyrightyear{2027}
\acmYear{2027}
\setcopyright{cc}
\setcctype{by}
\acmBooktitle{32nd Asia and South Pacific Design Automation Conference (ASPDAC '27), January 25--28, 2027, Tokyo, Japan}
\acmDOI{10.1145/3840404.3848321}
\acmConference[ASPDAC '27]{32nd Asia and South Pacific Design Automation Conference}{January 25--28, 2027}{Tokyo, Japan}
\acmISBN{979-8-4007-2923-2/2027/01}

\begin{document}

\title{HGRP: Heterogeneous Graph Neural Network for Routability Prediction}

\author{Yule Sheng}
\affiliation{%
  \institution{Nanjing University of Science and Technology}
  \city{Nanjing}
  \country{China}
}
\email{124132011758@njust.edu.cn}

\author{Yue Yuan}
\affiliation{%
  \institution{Nanjing University of Science and Technology}
  \city{Nanjing}
  \country{China}
}
\email{yuanyue1801@njust.edu.cn}

\author{Shan Shen}
\affiliation{%
  \institution{Nanjing University of Science and Technology}
  \city{Nanjing}
  \country{China}
}
\email{shanshen@njust.edu.cn}

\author{Daying Sun}
\affiliation{%
  \institution{Nanjing University of Science and Technology}
  \city{Nanjing}
  \country{China}
}
\email{hasdysun@njust.edu.cn}

\thanks{
This work was supported by the Fundamental Research Funds for the Central Universities (Grant No. 30925010605, 30924012004), Jiangsu Provincial Major Science and Technology Project (Grant No. BG2025012).
$^*$Corresponding author: S. Shen, D. Y. Sun.
}
%% \renewcommand{\shortauthors}{Anonymous et al.}

\begin{abstract}
  Routability prediction enables rapid evaluation of placement quality without routing, providing effective feedback for routability-driven placement optimization. Recent machine learning methods have achieved efficient routability prediction through data-driven approach. However, their prediction performance highly depends on the capability of input representations to capture essential circuit information. To address the limitations of existing circuit representations in simultaneously preserving logical topology, physical layout and routing resource information, this paper proposes HGRP, a heterogeneous graph neural network for routability prediction. HGRP models the cells, nets, and routing grids as distinct types of nodes and explicitly represents topological connectivity, spatial mapping, and grid adjacency, thereby constructing a heterogeneous graph representation that preserves multi-level circuit semantics. Based on the representation, a heterogeneous message passing mechanism is developed to enable information interaction and feature fusion among distinct types of nodes, effectively learning the complex correlations between circuit structures and routing behaviors. Experimental results on the CircuitNet dataset show that HGRP improves F1-score by 18.69\% relative improvements compared with the state-of-the-art in the congestion prediction task. The source code is publicly available at \url{https://github.com/ShenShan123/HGPR}.
\end{abstract}

%    <concept>
%        <concept_id>10010583.10010682.10010697</concept_id>
%        <concept_desc>Hardware~Physical design (EDA)</concept_desc>
%        <concept_significance>500</concept_significance>
%        </concept>
%  </ccs2012>

% \ccsdesc[500]{Hardware~Physical design (EDA)}




\maketitle

\section{Introduction}
As one of the most complex and time-consuming stages in physical design, if routability is not adequately considered early in the design flow, routing may suffer from congestion, design rule violations (DRVs), timing violations, and signal integrity issues. In such cases, the design flow often needs to return to the placement or even floorplan stage to adjust cell locations and rerun the routing process. Therefore, accurate routability prediction is essential for improving design quality and reducing design iterations.

Early routability-driven placement methods commonly rely on global routing \cite{6504553} or trial global routing, where routing results are used as feedback to guide placement optimization. However, repeatedly invoking global router during placement optimization often introduces prohibitive computational overhead. In contrast, fast estimation methods based on net density, such as probabilistic congestion estimation and RUDY \cite{4211973}, can substantially reduce the cost. Nevertheless, these methods lack explicit modeling of routing-path selection and competition for routing resource, making it difficult to accurately characterize actual local congestion.

Recent advances in machine learning provide solutions for rapid and accurate routability prediction, as large-scale EDA datasets \cite{10158384} \cite{r2g2026} have become increasingly available for model training and evaluation. Due to the inherent two dimensional spatial structure of physical layout, a typical approach is to formulate routability prediction as an image regression or image segmentation task. Methods such as RouteNet \cite{8587655}, J-Net \cite{10.1145/3372780.3375560}, GPDL \cite{9473959} and Pros \cite{9256565} encode physical design information into multichannel feature maps and apply convolutional neural networks (CNNs) to predict routability at grid-level, thereby providing rapid feedback for routability-driven design. However, modeling circuits as images may weaken or even discard topological information, potentially limiting generalization across netlists with different connectivity structures.

Graph neural networks (GNNs) have been introduced into routability prediction to improve the representation of topological information. Since very large scale integrated (VLSI) circuits can be naturally represented as a graph, methods such as CongestionNet \cite{8920342}, NetlistGNN \cite{3600270.3601747}, HybridNet \cite{DBLP:journals/corr/abs-2305-05374}, and LHNN \cite{10.1145/3489517.3530675} learn circuit representations via GNNs to predict routability metrics such as congestion and DRVs. Graph-based approaches generally model circuits at the cell, net, or grid-level. However, existing methods often emphasize net connectivity to incorporate topological information and perform prediction at either the cell-level or the grid-level, while overlooking the influence of the spatial mapping between cells and routing grids on routability. The absence of cell-level weakens the representation of cell-level structural information, whereas the absence of grid-level limits the capability to characterize routing resource competition and local congestion propagation. Consequently, both types of graph constructions suffer from incomplete semantic representation.

To address these limitations, we propose HGRP, a \underline{H}eterogeneous \underline{G}raph neural network for \underline{R}outability \underline{P}rediction. HGRP represents circuits as heterogeneous graphs, and introduces a heterogeneous message passing mechanism coupled with an adaptive feature aggregation strategy to update node representations. In this way, the model captures the complex coupling among circuit netlist, physical layout and routing resource. Moreover, the message passing order can be flexibly adjusted according to the dependency of different tasks on node representations, enabling HGRP to adapt to various physical design tasks beyond congestion prediction. The main contributions of this work are summarized as follows:
\begin{itemize}
\item We propose a \textit{heterogeneous graph representation} for routability prediction, in which cells, nets, and routing grids are modeled as distinct types of nodes. Topological connectivity, spatial mapping, and grid adjacency relations are explicitly encoded, providing a more complete circuit representation for routability prediction.
\item We develop a \textit{heterogeneous message passing mechanism} that jointly learns routing demands induced by netlist topology and local congestion characteristics constrained by physical layout.
\item We introduce an \textit{attention-based adaptive feature aggregation strategy} that dynamically determines the contributions of neighboring nodes according to their node representations and edge attributes during message passing.
\end{itemize}

\section{Problem Formulation}
\subsection{Congestion Prediction}
Routing congestion is an important metric for evaluating the routability of placement solution. During global routing, the chip region is partitioned into a regular \(w \times h\) grid graph. Based on the netlist and cell placement, the global router estimates the routing paths of individual nets and assigns the routing demand to the corresponding grid regions. Congestion occurs when the demand within a grid exceeds its available routing capacity, potentially leading to increased routing detours, timing degradation, and even DRVs during detailed routing. Following the definition adopted in CircuitNet \cite{10158384}, the congestion of grid \(g_{i,j}\) is formulated as
\begin{equation}
    congestion_{i,j}=\frac{max(demand_{i,j}-capacity_{i,j},0)}{capacity_{i,j}}
\end{equation}
where \(demand_{i,j}\) and \(capacity_{i,j}\) are the routing demand and capacity of \(g_{i,j}\), respectively, \(congestion_{i,j}\) measures the degree of routing congestion in the corresponding region and is clipped to zero when routing demand does not exceed capacity. The congestion of all grids are arranged according to their spatial locations to form the complete congestion map \(\mathbf{Y}_{cong}\in\mathbb{R}^{w\times h}\), and congestion prediction aims to learn a parameterized mapping \(f_{\theta,cong}\) from circuit information \(\mathbf{X}\) to the congestion map.
\begin{equation}
    \mathbf{\hat{Y}}_{cong}=f_{\theta,cong}(\mathbf{X})\in\mathbb{R}^{w\times h}
\end{equation}

\subsection{Net Wirelength Prediction}
Net wirelength is an important measure of subsequent routing cost, and is closely associated with interconnect delay, power consumption and routing-resource utilization. During placement, metrics such as half-perimeter wirelength (HPWL) are commonly used to rapidly estimate net length. However, HPWL does not explicitly account for routing layers, obstacles, design rules, resource competition or congestion-induced detours, therefore may deviate from the actual wirelength in the final physical implementation.

Predicting the final wirelength before detailed routing can facilitate the early identification of nets likely to require long routes, providing realistic feedback for cell repositioning, wirelength optimization and timing closure. Here, we use the actual post-detailed-routing wirelength as the prediction target. For a net \(n_j\), the detailed router generates its final routing solution that satisfies connectivity and design-rule constraints
\begin{equation}
    \mathcal{R}_j=(\mathcal{S}_j,\mathcal{U}_j)
\end{equation}
where \(\mathcal{S}_j\) denotes the set of routing segments of net \(n_j\) across all metal layers, and \(\mathcal{U}_j\) denotes the set of vias connecting different routing layers. The actual post-detailed-routing wirelength of net \(n_j\) is formulated as
\begin{equation}
    wirelength_j=\sum_{s\in\mathcal{S}_j}\ell(s)
\end{equation}
where \(\ell(s)\) denotes the physical length of routed segment \(s\). Net wirelength prediction aims to learn a parameterized mapping \(f_{\theta,length}\) form circuit information \(\mathbf{X}\) to the actual post-detailed-routing wirelengths of all nets.
\begin{equation}
    \mathbf{\hat{Y}}_{length}=f_{\theta,length}(\mathbf{X})
\end{equation}

\subsection{Evaluation Metrics}
Since routability prediction is formulated as a grid-level congestion regression task in this work, we employ the normalized root mean square error (NRMSE) and structural similarity index measure (SSIM) to comprehensively evaluate model performance.

{\bfseries Definition 1 (NRMSE).} NRMSE measures the normalized deviation between the predicted and ground-truth congestion.
\begin{equation}
    \mathrm{NRMSE}=\frac{1}{y_{\max }-y_{\min }} \cdot \sqrt{\frac{\sum_{i=1}^{h} \sum_{j=1}^{w}\left(y_{i, j}-\hat{y}_{i, j}\right)^{2}}{h \times w}}
\end{equation}
where \(y_{i, j}\) and \(\hat{y}_{i, j}\) are the ground-truth and predicted congestion of grid \(g_{i,j}\), \(h\) and \(w\) are the height and width of the congestion map, \(y_{max}\) and \(y_{min}\) are the maximum and minimum ground-truth. The normalization eliminates the influence of scale differences across circuit designs, thereby enabling fair performance comparisons. A lower NRMSE indicates higher numerical prediction accuracy.

{\bfseries Definition 2 (SSIM). } SSIM measures the structural similarity between the predicted and ground-truth congestion maps.
\begin{equation}
    \mathrm{SSIM}=\frac{(2 \mu_{y} \mu_{\hat{y}}+C_{1})(2 \sigma_{y \hat{y}}+C_{2})}{(\mu_{y}^{2}+\mu_{\hat{y}}^{2}+C_{1})(\sigma_{y}^{2}+\sigma_{\hat{y}}^{2}+C_{2})}
\end{equation}
where \(\mu_{y}\) and \(\mu_{\hat{y}}\) are the local means of the ground-truth and predicted congestion maps, \(\sigma_{y}^{2}\) and \(\sigma_{\hat{y}}^{2}\) are the variances, \(\sigma_{y \hat{y}}\) is the covariance. SSIM evaluates the similarity between the two congestion maps in terms of local intensity, contrast, and spatial structure. A higher SSIM indicates greater structural similarity between the predicted and ground-truth congestion maps.

In addition, congestion labels exhibit a highly long-tailed distribution. Therefore, we set the classification threshold to the 90th percentile of the congestion distribution in the training set. Grid exceeding the threshold are labelled as congestion hotpots. Accuracy, F1-score, ROC-AUC, and PR-AUC are employed to evaluate the capability of the model to identify high-risk congested regions.

{\bfseries Definition 3 (Accuracy). }Accuracy measures the proportion of correctly classified grids among all evaluated grids.
\begin{equation}
    \mathrm{Accuracy}=\frac{TP+TN}{TP+TN+FP+FN}
\end{equation}
where \(TP\), \(TN\), \(FP\), \(FN\) are the numbers of true-positive, true-negative, false-positive, and false-negative grids, respectively.

{\bfseries Definition 4 (F1-score). }F1-score is the harmonic mean of Precision and Recall.
\begin{equation}
    \mathrm{Precision}=\frac{TP}{TP+FP}
\end{equation}
\begin{equation}
    \mathrm{Recall}=\frac{TP}{TP+FN}
\end{equation}
\begin{equation}
    \mathrm{F1\text{-}score}=2\times\frac{\mathrm{Precision}\times\mathrm{Recall}}{\mathrm{Precision}+\mathrm{Recall}}
\end{equation}
Precision measures the proportion of predicted hotspots that are truly congested, whereas Recall measures the proportion of actual congestion hotspots successfully identified by the model.

{\bfseries Definition 5 (ROC-AUC). }The receiver operating characteristic curve plots the true-positive rate against the false-positive rate across different decision thresholds, and ROC-AUC is the area under this curve.

{\bfseries Definition 6 (PR-AUC). }The precision–recall curve describes the trade-off between Precision and Recall across different decision thresholds, and PR-AUC is the area under this curve.

Accuracy measures overall classification correctness, whereas F1-score jointly considers the precision and recall of congestion hotpots detection. ROC-AUC evaluates discrimination between congested and non-congested region across decision thresholds. PR-AUC focuses on detection performance for the congestion class. Since congested regions are relatively rare, F1-score and PR-AUC are used as the primary metrics for evaluating hotspot detection performance.

For the wirelength prediction task, the model predicts the wirelength of individual nets rather than reconstructing a spatial congestion map. Therefore, we adopt MAE, RMSE, Pearson correlation, Spearman correlation, and Kendall correlation as the evaluation metrics.

\begin{figure}[t]
  \centering
 \includegraphics[width=\linewidth]{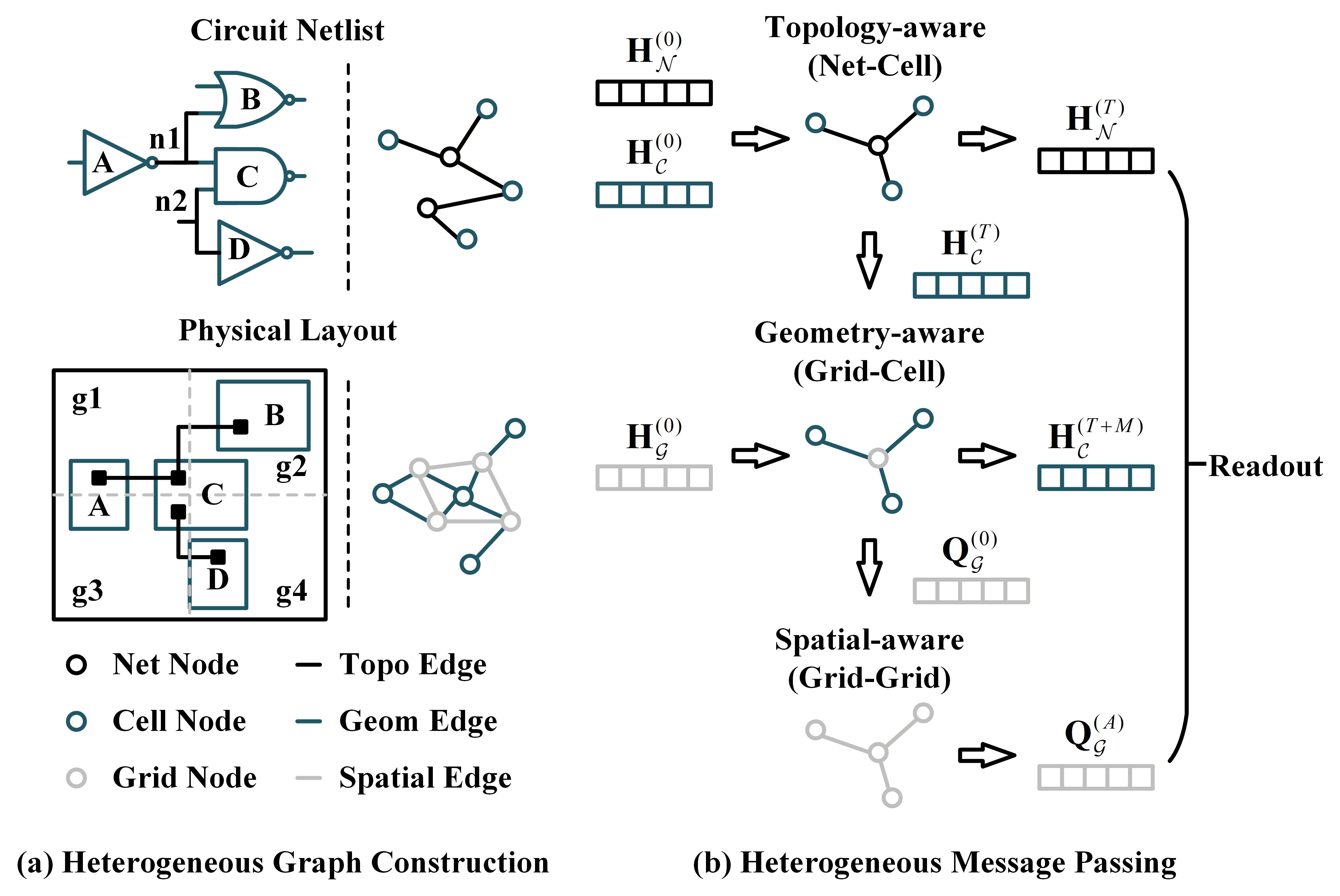}
  \caption{The HGRP architecture.}
  \Description{The heterogeneous graph and network architecture.}
\end{figure}

\section{Methodology}
\subsection{Graph Construction}
To preserve semantic information from logical topology, physical layout and routing resource, we represent each circuit as heterogeneous graph, as shown in Figure 1(a). HGRP models the cells, nets, and routing grids as distinct types of nodes, forming the node sets \(\mathcal{C}\), \(\mathcal{N}\) and \(\mathcal{G}\), respectively.Their input feature matrices are
\begin{equation}
    \mathbf{X_\mathcal{C}}\in\mathbb{R}^{|\mathcal{C}|\times d_\mathcal{C}},
    \mathbf{X_\mathcal{N}}\in\mathbb{R}^{|\mathcal{N}|\times d_\mathcal{N}},
    \mathbf{X_\mathcal{G}}\in\mathbb{R}^{|\mathcal{G}|\times d_\mathcal{G}}
\end{equation}
where \(d_{\mathcal{C}}\), \(d_{\mathcal{N}}\) and \(d_{\mathcal{G}}\) are the corresponding feature dimensions. HGRP defines three types of relation to encode topological connectivity, spatial mapping and grid adjacency
\begin{equation}
    \mathcal{T}\subseteq \mathcal{C}\times \mathcal{N},
    \mathcal{M}\subseteq \mathcal{C}\times \mathcal{G},
    \mathcal{A}\subseteq \mathcal{G}\times \mathcal{G}
\end{equation}
Specifically, \(\mathcal{T}\) connects cells to nets, \(\mathcal{M}\) connects cells to grids, and \(\mathcal{A}\) connects adjacent grids. To enable bidirectional information propagation across node types, HGRP introduces the reverse relations
\begin{equation}
    \mathcal{T}^{-1}=\left \{(n_j,c_i)|(c_i,n_j)\in\mathcal{T}\right \},
    \mathcal{M}^{-1}=\left \{(g_k,c_i)|(c_i,g_k)\in\mathcal{M}\right \}
\end{equation}
In addition, edge feature matrices are constructed for the spatial mapping, reverse spatial mapping, and grid adjacency relations
\begin{equation}
    \mathbf{Z_\mathcal{M}}\in\mathbb{R}^{|\mathcal{M}|\times d_\mathcal{M}},
    \mathbf{Z_{\mathcal{M}^{-1}}}\in\mathbb{R}^{|\mathcal{M}^{-1}|\times d_{\mathcal{M}^{-1}}},
    \mathbf{Z_\mathcal{A}}\in\mathbb{R}^{|\mathcal{A}|\times d_\mathcal{A}}
\end{equation}
providing fine-grained relational information for the subsequent attention-based adaptive message aggregation. Based on the above definitions, the heterogeneous graph representation of a circuit is formulated as
\begin{equation}
    \mathcal{H}=(\mathcal{V},\mathcal{E},\mathbf{X},\mathbf{Z})
\end{equation}
where \(\mathcal{V}=\mathcal{C}\cup \mathcal{N}\cup \mathcal{G}\) and \(\mathcal{E}=\mathcal{T}\cup \mathcal{T}^{-1}\cup \mathcal{M}\cup \mathcal{M}^{-1}\cup \mathcal{A}\) are the node and edge sets, \(\mathbf{X}\) and \(\mathbf{Z}\) are their feature matrices.

\subsection{Model Architecture}
The overall architecture of HGRP is illustrated in Figure 1(b). Given the heterogeneous graph representation \(\mathcal{H}=(\mathcal{V},\mathcal{E},\mathbf{X},\mathbf{Z})\), the raw features of different node and edge types may differ in both dimensionality and physical semantics. Therefore, HGRP employs independent linear projection layers to map the input features of each type into a unified latent representation space
\begin{equation}
    \mathbf{H}_t^{(0)}=\sigma(\mathbf{X}_t\mathbf{W}_t+\mathbf{b}_t),
    t\in\left \{\mathcal{C},\mathcal{N},\mathcal{G}\right \}
\end{equation}
\begin{equation}
    \mathbf{E}_r=\sigma(\mathbf{Z}_r\mathbf{W}_r^e+\mathbf{b}_r^e),
    r\in\left \{\mathcal{M},\mathcal{M}^{-1},\mathcal{A}\right \}
\end{equation}
where \(\mathbf{H}_t^{(0)}\) is the initial representation of nodes of type \(t\), \(\mathbf{E}_r\) is the representation of edges associated with relation \(r\), \(\sigma(\cdot)\) is a nonlinear activation function, \(\mathbf{W}_t\) and \(\mathbf{W}_r^e\), together with their corresponding biases, are learnable parameters. After feature encoding, the node representations are sequentially processed by topology-aware, geometry-aware, and spatial-aware message passing modules
\begin{equation}
    (\mathbf{H}_\mathcal{C}^{(T)},\mathbf{H}_\mathcal{N}^{(T)})=
    \Phi_{T}(\mathbf{H}_\mathcal{C}^{(0)},\mathbf{H}_\mathcal{N}^{(0)};
    \mathcal{T},\mathcal{T}^{-1})
\end{equation}
\begin{equation}
    (\mathbf{H}_\mathcal{C}^{(T+M)},\mathbf{H}_\mathcal{G}^{(M)})=
    \Phi_{M}(\mathbf{H}_\mathcal{C}^{(T)},\mathbf{H}_\mathcal{G}^{(0)};
    \mathcal{M},\mathcal{M}^{-1};\mathbf{E_\mathcal{M}},\mathbf{E_{\mathcal{M}^{-1}}})
\end{equation}
\begin{equation}
    \mathbf{Q}_\mathcal{G}^{(A)}=\Phi_{A}(\mathbf{Q}_\mathcal{G}^{(0)};\mathcal{A};\mathbf{E_\mathcal{A}}),
    \mathbf{Q}_\mathcal{G}^{(0)}=\sigma
    ([\mathbf{X}_\mathcal{G}||\mathbf{H}_\mathcal{G}^{(M)}]\mathbf{W}_f+\mathbf{b}_f)
\end{equation}
where \(||\) denotes feature concatenation, \(T\), \(M\), \(A\) are the numbers of layers in the corresponding modules. The topology-aware module \(\Phi_{T}\) incorporates contextual information from multi-pin nets into the cell representations. Then the geometry-aware module \(\Phi_{M}\) transfers cell-level topology and placement information to the corresponding grid. Finally, the spatial-aware module \(\Phi_{A}\) aggregates placement and routing resource information from adjacent grids, enlarging the spatial receptive field and capturing routing resource competition and spatial correlations among congested regions.

The grid representations before and after spatial propagation are concatenated and passed to a multi-layer perceptron (MLP) to predict grid-level congestion
\begin{equation}
    \mathbf{\hat{y}}_{cong}=\mathrm{MLP}[\mathbf{Q}_\mathcal{G}^{(0)}||\mathbf{Q}_\mathcal{G}^{(A)}]
    \in\mathbb{R}^{|\mathcal{G}|}
\end{equation}
Then the predictions are rearranged according to the physical coordinates, forming the congestion map \(\mathbf{\hat{Y}}_{cong}\in\mathbb{R}^{w\times h}\).

\subsection{Heterogeneous Message Passing}
HGRP employs a staged message passing mechanism according to the physical semantics of each edge relation as shown in Figure 1(b). Node representations propagate sequentially along the cell-net connection, cell-grid mapping and grid adjacency. For any relation \(r\), a target node \(v_i\) aggregates information from its relation-specific neighbourhood \(N_r(v_i)\)
\begin{equation}
    \mathbf{m}_{i,r}^{(l)}=\mathrm{AGG}_{v_j\in N_r(v_i)}
    \mathcal{F}_r^{(l)}(h_i^{(l)},h_j^{(l)},e_{ji,r})
\end{equation}
where \(\mathcal{F}{r}^{(l)}(\cdot)\) is the message generation function associated with relation \(r\), \(\mathrm{AGG}(\cdot)\) is the neighbourhood aggregation operator, \(e_{ji,r}\) is the attribute of edge \((v_j,v_i)\) under relation \(r\). The edge attribute term is omitted when the corresponding relation does not contain explicit edge features. The node representation is updated through a residual connection \cite{7780459} and layer normalization
\begin{equation}
    \mathbf{h}_i^{{(l+1)}}=\mathrm{LN}(\mathbf{h}_i^{(l)}+
    \sigma(\mathbf{W}_r^{o,(l)}\mathbf{m}_{i,r}^{(l)}+\mathbf{b}_r^{o,(l)}))
\end{equation}
where \(\mathrm{LN}(\cdot)\) is layer normalization. The residual connection preserves the previous node representation and improves training stability across multiple message passing layers. All message passing stages follow the same node update formulation, while adopting relation-specific message generation and neighbourhood aggregation operations.

{\bfseries Topology-aware Message Passing: }The topology-aware stage performs bidirectional message passing along \(\mathcal{T}\) and \(\mathcal{T}^{-1}\). The neighbourhood messages for net node \(n_j\) and cell node \(c_i\) are formulated as
\begin{equation}
    \mathbf{m}_{n_j}^{(l)}=\mathbf{W}_{\mathcal{T},s}^{(l)}\mathbf{h}_{n_j}^{(l)}+
    \mathbf{W}_{\mathcal{T},n}^{(l)}
    \sum_{c_i\in N_{\mathcal{T}}(n_j)}\mathbf{h}_{c_i}^{(l)}
\end{equation}
\begin{equation}
    \mathbf{m}_{c_i}^{(l)}=\mathbf{W}_{\mathcal{T}^{-1},s}^{(l)}\mathbf{h}_{c_i}^{(l)}+
    \mathbf{W}_{\mathcal{T}^{-1},n}^{(l)}
    \sum_{n_j\in N_{\mathcal{T}^{-1}}(c_i)}\mathbf{h}_{n_j}^{(l)}
\end{equation}
In the cell-to-net direction, each net node aggregates its incident cell features to capture multi-pin net context. The aggregated net information is then propagated back to connected cells, enabling each cell to incorporate information from other cells connected through the same net. Through multiple bidirectional propagation layers, cell representations encode both intrinsic attributes and topological context.

{\bfseries Geometry-aware Message Passing: }The updated cell representations \(\mathbf{H}_\mathcal{C}^{(T)}\) and the initial grid representations \(\mathbf{H}_\mathcal{G}^{(0)}\) are passed to the geometry-aware stage, where bidirectional message passing is performed along \(\mathcal{M}\) and \(\mathcal{M}^{-1}\). An edge-aware multi-head attention mechanism \cite{vaswani2023attentionneed} is introduced to dynamically adjust the contributions of neighboring nodes. For relation \(r\in{\mathcal{M},\mathcal{M}^{-1}}\), the attention weight from source node \(v_j\) to target node \(v_i\) under head \(p\) is formulated as
\begin{equation}
    \alpha_{ji,r}^p=\mathrm{softmax}_{v_j\in N_r(v_i)}
    (\frac{(\mathbf{W}_{Q, r}^{p} \mathbf{h}_{i})^\mathrm{T}
    (\mathbf{W}_{K, r}^{p} \mathbf{h}_{j}+\mathbf{W}_{E, r}^{p}\mathbf{e}_{j i, r})}
    {\sqrt{d_p}})
\end{equation}
where \(d_p\) is the feature dimension of each attention head. The message received by \(v_i\) under attention head (p) is
\begin{equation}
    \mathbf{m}_{i,r}^p=\sum_{v_j\in N_r(v_i)}\alpha_{ji,r}^p
    (\mathbf{W}_{V, r}^{p} \mathbf{h}_{j}+\mathbf{W}_{E, r}^{p}\mathbf{e}_{j i, r})
\end{equation}
Outputs from the \(P\) attention heads are concatenated to obtain the final aggregated message
\begin{equation}
    \mathbf{m}_{i,r}=||_{p=1}^P\mathbf{m}_{i,r}^p
\end{equation}
In the cell-to-grid direction, each grid node aggregates information from its spatially mapped cells, thereby transferring cell-level topology and placement features to the corresponding physical region. In the reverse direction, each cell aggregates information from its associated grid region, allowing its representation to capture local routing resource conditions.

\begin{table*}[t]
  \centering
  \caption{Congestion Prediction Results}
  \label{tab:congestion}
  \begin{tabular}{ccccccccccc}
    \toprule
    Model & NRMSE$\downarrow$ & SSIM$\uparrow$ & Accuracy$\uparrow$ & F1-score$\uparrow$ & ROC-AUC$\uparrow$ & PR-AUC$\uparrow$ & Param & Memory & Train & Inference \\
    \midrule
    GPDL       & 0.0495 & 0.7789 & 0.8682          & 0.3558 & 0.7773 & 0.4055 & 0.119M & 0.309G & 8.100s & 0.415s \\
    NetlistGNN & 0.0406 & 0.7511 & \textbf{0.8966} & 0.3992 & 0.8699 & 0.4118 & 0.271M & 5.310G & 33.678s & 4.279s \\
    LHNN       & 0.0350 & 0.7750 & 0.8830          & 0.4589 & 0.8658 & 0.4731 & 0.054M & 3.486G & 28.615s & 3.354s \\
    HGRP       & \textbf{0.0338} & \textbf{0.8090} & 0.8803
               & \textbf{0.5447} & \textbf{0.8998} & \textbf{0.5428} & 0.200M & 9.495G & 60.356s & 8.542s \\
    \bottomrule
  \end{tabular}
\end{table*}

\begin{table*}[t]
  \centering
  \caption{Multi-seed Experimental Results.}
  \label{tab:multi-seed}
  \begin{tabular}{ccccccc}
    \toprule
    Random Seed & NRMSE$\downarrow$ & SSIM$\uparrow$ & Accuracy$\uparrow$ & F1-score$\uparrow$ & ROC-AUC$\uparrow$ & PR-AUC$\uparrow$ \\
    \midrule
    1            & 0.0338 & 0.8083 & 0.8847 & 0.5360 & 0.8994 & 0.5084 \\
    2            & 0.0339 & 0.8061 & 0.8801 & 0.5470 & 0.8807 & 0.5313 \\
    3            & 0.0338 & 0.8088 & 0.8844 & 0.5313 & 0.9026 & 0.5095 \\
    4            & 0.0338 & 0.8092 & 0.8821 & 0.5325 & 0.9013 & 0.5174 \\            
    \midrule
    Mean$\pm$Std & 0.0338$\pm$0.0000 & 0.8081$\pm$0.0014 & 0.8828$\pm$0.0021 & 0.5367$\pm$0.0071 & 0.8960$\pm$0.0102 & 0.5167$\pm$0.0105 \\
    \bottomrule
  \end{tabular}
\end{table*}

{\bfseries Spatial-aware Message Passing: }The grid representations obtained from the geometry-aware stage are concatenated with the original grid features, thereby preserving both routing resource information and higher-level semantics aggregated from the cell nodes. Then HGRP performs spatial message passing over the grid-adjacency relation \(\mathcal{A}\) to model routing resource competition and congestion propagation between neighbouring regions.

Let \(\mathbf{h}_i\) and \(\mathbf{h}_j\) denote the fused representations of target grid \(g_i\) and adjacent grid \(g_j\), respectively. The spatial attention weight under attention head (p) is formulated as
\begin{equation}
    \alpha_{ji}^{p}=\mathrm{softmax}_{g_j\in N_\mathcal{A}(g_i)}
    (\frac{(\mathbf{W}_{Q}^{p}\mathbf{h}_i)^\mathrm{T}
    (\mathbf{W}_{K}^{p}\mathbf{h}_j+
    \mathbf{W}_{E}^{p}\mathbf{e}_{ji,\mathcal{A}}))}{\sqrt{d_p}})
\end{equation}
The messages from adjacent grids are aggregated across \(P\) attention heads as
\begin{equation}
    \mathbf{m}_i=||_{p=1}^P\sum_{g_j\in N_\mathcal{A}(g_i)}\alpha_{ji}^p
    (\mathbf{W}_{V}^{p}\mathbf{h}_j+\mathbf{W}_{E}^{p}\mathbf{e}_{ji,\mathcal{A}})
\end{equation}
As spatial message-passing layers are stacked, each grid node progressively receives placement density and routing resource information from its multi-hop neighborhood. The resulting spatial context captures the local continuity of congested regions and improves the representation of routing resource competition and congestion propagation.

\begin{table*}[t]
  \centering
  \caption{Net Wirelength Prediction Results.}
  \label{tab:wirelength}
  \begin{tabular}{cccccc}
    \toprule
    Model & MAE$\downarrow$ & RMSE$\downarrow$ 
          & Pearson$\uparrow$ & Spearman$\uparrow$ & Kendall$\uparrow$ \\
    \midrule
    NetlistGNN & 0.2945 & 0.5652 & 0.6621 & 0.7109 & 0.5352 \\
    LHNN       & 0.3157 & 0.5587 & 0.6214 & 0.6966 & 0.5270 \\
    HGRP       & \textbf{0.2925} & \textbf{0.5138}
               & \textbf{0.6932} & \textbf{0.7354}
               & \textbf{0.5636} \\
    \bottomrule
  \end{tabular}
\end{table*}

\begin{table*}[t]
  \centering
  \caption{Ablation Study Results.}
  \label{tab:ablation}
  \begin{tabular}{ccccccc}
    \toprule
    Variant & NRMSE$\downarrow$ & SSIM$\uparrow$ & Accuracy$\uparrow$ & F1-score$\uparrow$ & ROC-AUC$\uparrow$ & PR-AUC$\uparrow$ \\
    \midrule
    W/o grid      & 0.0392 & 0.7594 & 0.9117 & 0.3810 & 0.8011 & 0.3178 \\
    W/o attention & 0.0340 & 0.8039 & \textbf{0.9172} & 0.4577 & 0.8698 & 0.4592 \\
    Mixed         & 0.0340 & 0.8076 & 0.9147 & 0.4533 & 0.8783 & 0.4800 \\
    HGRP          & \textbf{0.0338} & \textbf{0.8090} & 0.8803
                  & \textbf{0.5447} & \textbf{0.8998}
                  & \textbf{0.5428} \\
    \bottomrule
  \end{tabular}
\end{table*}

\begin{table*}[t]
  \centering
  \caption{Performance Comparison under Different Hyperparameter Settings.}
  \label{tab:hyperparameter}
  \begin{tabular}{ccccccccc}
    \toprule
    Node Dimension & Edge Dimension & Layers & NRMSE$\downarrow$ & SSIM$\uparrow$ & Accuracy$\uparrow$ & F1-score$\uparrow$ & ROC-AUC$\uparrow$ & PR-AUC$\uparrow$ \\
    \midrule
    32  & 16 & 2 & 0.0337 & 0.8082 & 0.9055 & 0.4846 & 0.8569 & 0.4825 \\
    64  & 16 & 2 & 0.0338 & 0.8090 & 0.8803 & \textbf{0.5447} & 0.8998 & \textbf{0.5428} \\
    128 & 16 & 2 & \textbf{0.0336} & 0.8059 & 0.9211 & 0.4688 & 0.8789 & 0.4562 \\
    64  & 32 & 2 & 0.0337 & 0.8083 & 0.9194 & 0.4388 & 0.8743 & 0.4460 \\
    64  & 16 & 3 & 0.0336 & \textbf{0.8104} & \textbf{0.9216} & 0.4522 & \textbf{0.9022} & 0.4565 \\
    \bottomrule
  \end{tabular}
\end{table*}

\section{Experiments}
\subsection{Experimental Settings}
We conduct experiments on the CircuitNet \cite{10158384} dataset to evaluate the effectiveness of HGRP for routability prediction. Following the experimental settings of CircuitNet, RISCY-a, RISCY-b, RISCY-FPU-a, and RISCY-FPU-b are used for training, while zero-riscy-a and zero-riscy-b are used for validation and testing. This split is intended to demonstrate the generalization capability of the model outside the training dataset. All samples utilize the routability features and routing congestion labels provided by CircuitNet. The extracted routability features consist of topological features from circuit connectivity, geometric features from physical layout, and local statistical features at the grid level.

We implement HGRP using PyTorch Geometric with an NVIDIA A100 GPU. The hidden dimensions of cell, net, and grid node representations are uniformly set to 64, while the hidden dimension of edge features is set to 16. The topology-aware, geometry-aware, and spatial-aware stages each consist of two message passing layers. We used AdamW\cite{loshchilov2019iclr-decoupled} optimizer with initial learning rate of \(5e^{-4}\) and MSE loss to train HGRP.

\subsection{Experimental Results}
We compared HGRP with GPDL \cite{9473959}, NetlistGNN \cite{3600270.3601747} and LHNN \cite{10.1145/3489517.3530675} to evaluate its predictive performance. These methods adopt distinct circuit representations, enabling the effectiveness of the proposed heterogeneous circuit representation to be evaluated from multiple perspectives.

GPDL encodes physical layout information into multichannel feature maps and resizes feature maps and congestion maps of different dimensions to \(256\times256\) through interpolation. NetlistGNN models cells and nets as heterogeneous nodes and introduces edges between spatially adjacent cells using a sliding window, thereby incorporating geometric information into the topology-based graph representation. The geometric window size is set to \(2.25\times2.25\), matching the routing grid dimensions. LHNN represents each circuit as a lattice hypergraph, allowing congestion information to propagate through both geometric and topological relations. In our implementation, the network architectures and unspecified parameters followed the corresponding publications or their public implementations. All methods were retrained using above dataset partition and supervised with routing congestion labels from the same source. Predictions from each method were arranged into the corresponding grid-level congestion maps before evaluation.

{\bfseries Congestion Prediction: }The experimental results for congestion prediction are reported in Table 1, HGRP achieves the best performance in terms of both NRMSE and SSIM, demonstrating that HGRP can reduce the overall prediction error of grid-level congestion while accurately reconstructing the spatial structure of congested regions.

The congestion labels in CircuitNet exhibit substantial class imbalance, with non-congested grids substantially outnumbering congested grids. Accuracy measures overall classification correctness but is easily dominated by the majority class, thus failing to adequately reflect the capability of model to identify minority congestion hotpots. In contrast, F1-score jointly measures the precision and recall of congestion hotpots, whereas PR-AUC evaluates performance across the precision-recall trade-off. These metrics therefore provided more informative assessments of minority hotspot detection. Compared with the best-performing baseline, HGRP improves F1-score and PR-AUC by 18.69\% and 14.73\%, respectively, demonstrating its effectiveness in identifying high-risk congested regions.

To evaluate the robustness of HGRP against random initialization, we conduct experiments with four different random seeds. As shown in Table 2, the regression metrics remain highly stable with negligible variations, while the hotspot prediction metrics exhibit slightly larger fluctuations due to their sensitivity to classification thresholds. Nevertheless, HGRP consistently maintains competitive performance across all runs, demonstrating its robustness.

{\bfseries Net Wirelength Prediction: }We further evaluated the applicability of the proposed heterogeneous graph representation through net wirelength prediction. Unlike congestion prediction, wirelength is predicted from net node representations. Therefore, we place geometry-aware stage before topology-aware stage. Cell nodes first incorporate layout information from their corresponding physical regions. Then net nodes aggregated the combined topological and geometric representations of their incident cells for wirelength regression.. 

The experimental results for net wirelength prediction are reported in Table 3, HGRP achieves the best performance across all evaluation metrics, demonstrating that HGRP can estimate net wirelength more accurately while maintaining higher consistency with the ground-truth wirelengths in terms of both numerical variation and net ranking.

These results demonstrate that the proposed heterogeneous graph representation effectively preserves both circuit netlist and physical layout information. Moreover, the propagation order of different relations can be dynamically adjusted according to the prediction target, enabling HGRP to accommodate physical design tasks at different abstraction levels.

\subsection{Ablation Experiment}
To evaluate the contributions of the heterogeneous graph representation and heterogeneous message-passing mechanism, we constructed three ablation variants of the complete HGRP model. W/o grid maps grid features to cell nodes before removing grid nodes and their associated relations. Cell representations are learned solely through the topological relations, after which the cell-level predictions are projected onto the grid graph. W/o attention replaces the attention-based adaptive aggregation in the geometry-aware and spatial-aware stages with equal-weight neighborhood aggregation. Mixed jointly aggregates information from different relation-specific neighborhoods rather than propagating sequentially through the topology-aware, geometry-aware, and spatial-aware stages.

The ablation study results are reported in Table 4. Removing grid nodes produces substantial decreases across all key metrics. Cell-level representations alone are therefore insufficient to capture the spatial distribution of routing demand. In contrast, explicitly introducing grid nodes enables the model to integrate logical topology, physical layout, and routing resource information at the spatial resolution consistent with the prediction target. Removing the attention mechanism reduced both F1-score and PR-AUC, demonstrating that adaptively adjusting neighbor contributions based on node representations and edge attributes improves the model's capability to identify congestion hotspots. Mixed message passing also reduced predictive performance. Jointly aggregating different relation types may blur their distinct semantics, whereas staged propagation preserves relation-specific information and captures dependencies across circuit design hierarchies.

\subsection{Hyperparameter Analysis}
To analyze the impact of architectural hyperparameters on prediction performance, we compared HGRP under different numbers of node hidden dimensions, edge hidden dimensions, and message-passing layers, as reported in Table 5. Regression metrics remained relatively stable across these settings, whereas congestion hotspot detection metrics varied more substantially. This observation suggests that model capacity primarily affects the discrimination of the minority congested samples, with relatively limited influence on fitting the overall congestion distribution.

As the node hidden dimension increases from 32 to 64, the additional capacity may facilitate representation of interactions among logical topology, physical layout and routing resource, which contributes to the overall improvement in classification metrics. However, further increasing the dimension to 128 markedly reduces F1-score and PR-AUC. A possible explanation is that the increased number of model parameters produced more conservative predictions, favouring the majority low-congestion class while reducing recall for minority congestion hotspots.

Similarly, increasing the edge hidden dimension does not produce consistent performance improvements. Since spatial-mapping and grid-adjacency relations have relatively explicit physical semantics, low-dimensional edge representations may already capture the principal spatial information. Further increasing the dimension may introduce redundant features that interfere with attention learning, making neighbour importance more difficult to distinguish.

Increasing the message-passing depth expands the receptive field, allowing grid nodes to aggregate topological and spatial information from larger neighbourhoods. Nevertheless, excessive propagation depth may cause neighbouring node representations increasingly similar, weakening the feature distinction between high-congestion grids and their surroundings, thereby reducing hotspot-classification performance.

In summary, moderate node and edge representation dimensions, together with appropriate message-passing depth, provide a better balance among representational capacity, training stability and generalization performance.

\section{Conclusion}
In this work, we propose HGRP, which models circuits as heterogeneous graphs to provide a more complete circuit representation for routability prediction. Based on the heterogeneous graph, we introduces a heterogeneous message passing mechanism coupled with an adaptive feature aggregation strategy to update node representations, thereby capturing the complex coupling among circuit netlist, physical layout and routing resource. Experimental results on the CircuitNet dataset evaluate the effectiveness of the the proposed method. In future work, we will further explore its application to routability-driven placement optimization.

\bibliographystyle{ACM-Reference-Format}
\bibliography{sample-base}

\end{document}